\documentclass{bmvc2k}

\usepackage{times}
\usepackage{epsfig}
\usepackage{graphicx}
\usepackage{amsmath}
\usepackage{amssymb}
\usepackage{multirow}
\usepackage{bbding}

\title{Uncertainty-aware Human Motion Prediction}

\addauthor{Pengxiang Ding}{}{1}
\addauthor{Jianqin Yin}{}{1}

\addinstitution{}
\addinstitution{}

\runninghead{}{}

\begin{document}

\maketitle

\begin{abstract}

% Version2
Human motion prediction is essential for tasks such as human motion analysis and human-robot interactions. Most existing approaches have been proposed to realize motion prediction. However, they ignore an important task, the evaluation of the quality of the predicted result. It is far more enough for current approaches in actual scenarios because people can't know how to interact with the machine without the evaluation of prediction, and unreliable predictions may mislead the machine to harm the human. Hence, we propose an uncertainty-aware framework for human motion prediction (UA-HMP). Concretely, we first design an uncertainty-aware predictor through Gaussian modeling to achieve the value and the uncertainty of predicted motion. Then, an uncertainty-guided learning scheme is proposed to quantitate the uncertainty and reduce the negative effect of the noisy samples during optimization for better performance. Our proposed framework is easily combined with current SOTA baselines to overcome their weakness in uncertainty modeling with slight parameters increment. Extensive experiments also show that they can achieve better performance in both short and long-term predictions in H3.6M, CMU-Mocap.

\end{abstract}

%-------------------------------------------------------------------------
\section{Introduction}

Human motion prediction aims to generate future skeleton sequences according to past observed ones. This technique can help machines anticipate human motion in the future and conjecture the intention of human action. Therefore, human motion prediction is essential to facilitate tasks such as human action analysis and human-robot interaction.

Many works have been proposed to improve the accuracy of human motion prediction in recent years. They are mainly divided into two types: methods based on sequential networks (RNNs) and methods based on feedforward networks (CNNs and GCNs). As to those methods based on RNNs\cite{16,17,18,21,22}, they can exploit rich temporal correlation of given human motion sequences due to their superior ability in sequence modeling. However, it is hard for them to make use of the spatial structure of skeleton because sequential modeling is only utilized on the time dimension. Therefore, CNNs\cite{19,20} and GCNs\cite{23,34,33,32,38,35} are more widely used recently. They can model spatiotemporal features of human motion simultaneously and thus boost the performance significantly.

However, there still exists one universe problem in most of the current methods. The evaluation of the quality of the predicted result, which is vital in actual scenarios, is neglected. In the test phase, they all rely on the pre-trained model to offer final prediction results. Thus, the predicted value is determinate and unique, which is far more enough in the real physical system involving human beings, like human-robot reactions. If the people can’t evaluate the predicted result, they can’t  know how to interact with the machine. Besides, the wrong or unreliable prediction result may mislead the machine to harm the people. Therefore, it is vital to assess the quality of prediction.

Therefore, in this paper, we present our uncertainty-aware framework for human motion prediction (UA-HMP). It mainly includes two core components. First, we design an uncertainty-aware predictor through Gaussian modeling, where the predicted joints’ coordinates of motion are modeled as the Gaussian parameters (i.e., the mean and variance). In this way, we can utilize determinative values to estimate the uncertainty of predicted joints coordinates. Second, we present an uncertainty-guided learning scheme to quantitate the uncertainty and achieve better model convergence. In particular, the noisy samples with high uncertainty are penalized during optimization to reduce their negative effect. Notably, our proposed framework can be easily combined with any current baselines to overcome their weakness in uncertainty modeling with slight parameters increment. As a result, our proposed framework has significant practical value for its great generalization on current baselines.

The main contributions of this paper are summarized as follows.

\begin{itemize}

\item We first propose an end-to-end learning framework human motion prediction(UA-HMP) to model the uncertainty in human motion prediction.

\item We present an uncertainty-aware predictor and uncertainty-guided learning scheme, where the former can get the value and the uncertainty of predicted motion through Gaussian modeling simultaneously, and the latter can quantitate the uncertainty and achieve better model convergence by penalizing the noisy samples during optimization.

\item Our proposed framework has significant practical value due to its great generalization on current baselines with slight parameters increment.

\end{itemize}

%-------------------------------------------------------------------------
\section{Related Work}

\subsection{Human motion prediction}

Skeleton-based motion prediction has attracted increasing attention recently. Recent works using neural networks \cite{08,16,17,18,19,20,21,22,23,34,33,32,38,35} have significantly outperformed traditional approaches \cite{12,13}.

RNNs\cite{16,17,18} were first used to predict human motion for their ability on sequence modeling. The first attempt was made by Fragkiadaki et al. [3], who proposed an Encoder-Recurrent-Decoder (ERD) model to combine encoder and decoder with recurrent layers. They encoded the skeleton in each frame to a feature vector and built temporal correlation recursively. Julieta et al.\cite{17} introduced a residual architecture to predict velocities and achieved better performance. However, these works all suffer from discontinuities between the observed poses and the predicted future ones. Though Gui et al.\cite{18} proposed to generate a smooth and realistic sequence through adversarial training, it is hard to alleviate error-accumulation in a long-time horizon inherent to the RNNs scheme. 

Recently, feedforward networks were widely adopted to help alleviate those above questions because their prediction is not recursive and thus could avoid error accumulation. Li et al.\cite{19} introduced a convolutional sequence-to-sequence model that encodes the skeleton sequence as a matrix whose columns represent the pose at every temporal step. However, their spatiotemporal modeling is still limited by the convolutional filters’ size. Recently, \cite{08,20} were proposed to consider global spatial and temporal features simultaneously. They all transformed temporal space to trajectory space and take the global temporal information into account. It contributes to capturing richer temporal correlation and thus achieved state-of-the-art results. Besides, Cai et al. \cite{32} also introduced transformer structure into this domain. To test the effectiveness and generalization of our method, in this paper, we choose the \cite{08} as a baseline of methods based on GCNs and \cite{20} as a baseline of methods based on CNNs to combine with our proposed framework.

\subsection{Uncertainty Measurement}

Despite the heavy success in deep learning, practical methods for estimating uncertainties in the predictions with deep networks have only recently become actively studied. Ghahramani et al.\cite{41} first propose to estimate the predictive variance of a deep neural network by computing the mean and variance of the sample, which is later referred to as the Monte Carlo (MC) dropout. Different from \cite{41} using a standard neural network, \cite{43} used a density network whose output consists of both mean and variance of a prediction trained with a negative log-likelihood criterion. Kendall et al.\cite{40} decomposed the predictive uncertainty into two major types and used a slightly different cost function for numerical stability. 

Although the general uncertainty-aware methods have been widely concerned in the NN-based applications, such as medical image segmentation and video segmentation, there is little research on the uncertainty of motion prediction. In methods\cite{44,45}, online adaptation methods were utilized for uncertainty estimation of human motion. There are two main drawbacks of them. First, their frameworks are not end-to-end, which makes the whole process complicated in actual scenarios. Second, they are not trainable, which makes the predicted result unreliable because they can’t use the information of training samples. To our best knowledge, this paper is the first to introduce an end-to-end framework to measure the uncertainty of predicted motion.

\section{Method}
In this section, we first formulate human motion prediction(HMP) with a brief illustration. Next, we will demonstrate two core components of our uncertainty-aware framework for human motion prediction (UA-HMP): an uncertainty-aware predictor and an uncertainty-guided learning scheme. The former utilizes Gaussian modeling to generate the value and uncertainty of the prediction. The latter aims to quantitate the uncertainty and achieve better model convergence by penalizing the noisy samples during optimization.

\begin{figure}[h]

\centering 
\includegraphics[width=5in]{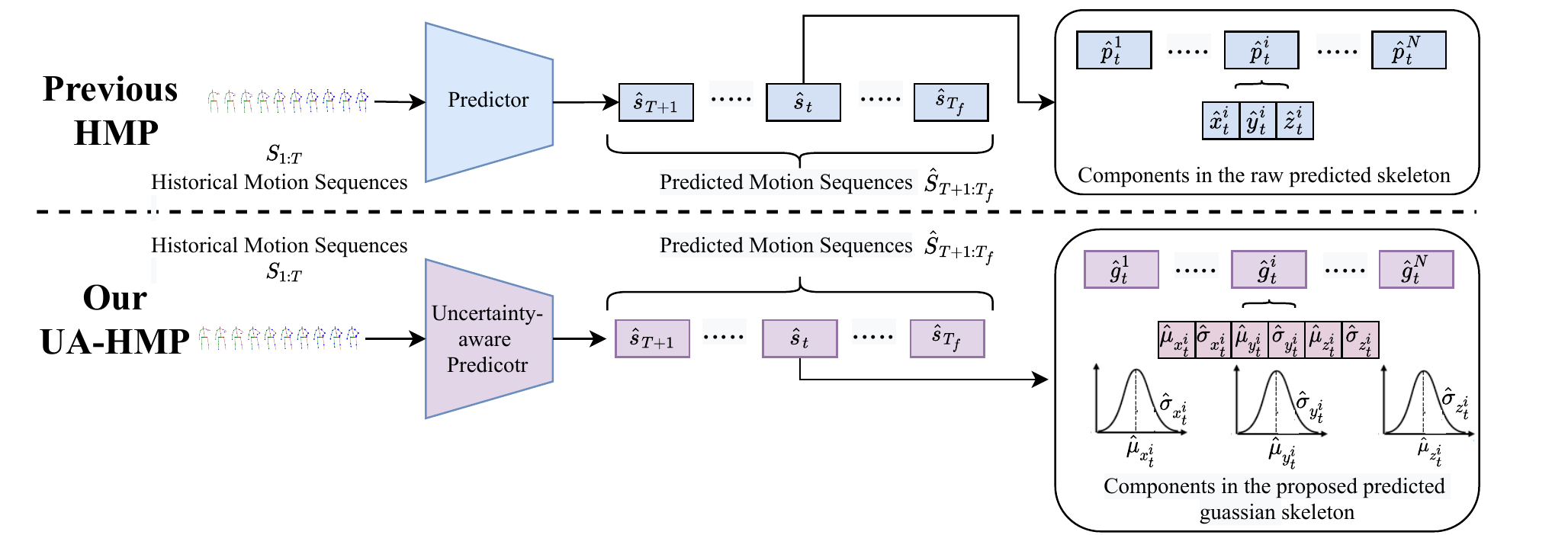} 

\caption{Overview of previous HMP and our UA-HMP. The top is the overall process of previous methods of HMP and the bottom is our proposed UA-HMP. Given the historical motion sequence $S_{1:T}$, the main differences of these two frameworks are the predicted parameters of the output. In HMP, the output are the predicted joint coordinates while in UA-HMP the output are a group of Gaussian parameters of predicted joint.}
\label{overall_framework}
\end{figure}

\subsection{Problem Formulation}
As shown in Figure \ref{overall_framework}, we illustrate the overall process of previuos HMP.  We denote the historical $3D$ skeleton-based poses as $S_{1:T}=\{s_t\}_{t=1}^T$ and future poses as $S_{T+1:T_f}=\{s_t\}_{t=T+1}^{T_f}$, where $s_t\in\mathbb{R}^{N\times D}$ represents the pose at frame $t$. For the task HMP, with the input of $S_{1:T}$, the goal is to generate predicted motion sequence ${\hat{S}}_{T+1:T_f}=\{\hat s_t\}_{t=T+1}^{T_f}$, where $\hat s_t\in\mathbb{R}^{N\times D}$ represents the predicted pose at frame $t$. Specifically, the groudtruth pose $s_t$ and predicted pose $\hat s_t$ at frame $t$ are composed of $N$ joint corrdinates. We here take $\hat s_t$ as an example: $\hat s_t=\{\hat p_t^i\}_{i=1}^N\in\mathbb{R}^{N\times D}$, where $\hat p_t^i=(\hat x_t^i,\hat y_t^i,\hat z_t^i)\in\mathbb{R}^{D}$ denotes the predicted $i$th joint coordinates in frame $t$ and the $D=3$ depicts the dimension of joint coordinates.

\subsection{Uncertainty-aware predictor}

In this section, we analyze the importance of uncertainty modeling firstly. Next, we explain why it is difficult for the previous HMP methods to model uncertainty. At last, we will demonstrate the overall process of the proposed UA-HMP.

First, we emphasize the importance of uncertainty modeling of human motion. There exists a universal problem in current approaches that they ignore the evaluation of the quality of the predicted result. Concretely, the information provided by current models only contains the positions of the future pose, and then the prediction confidence is unknown. This phenomenon is not terrible in scenarios like recommendation systems; however, it is dangerous in human-robot interactions because the wrong prediction may mislead machines to harm people. In brief, the uncertainty of human motion is vital in terms of security.

Then, we analyze the limitation of previous approaches to model uncertainty. There mainly exist two reasons. On the one hand, unlike the classification task, the predicted pose is output as deterministic coordinate values instead of a score. Thus, it can’t be used to measure the uncertainty of the predicted pose. On the other hand, because there is only a correct answer/label for the predicted pose, complex modeling is not required for predicting the uncertainty. In other words, if the correct answer/label is a distribution, the uncertainty of coordinates can be modeled by measuring the value of the variance of this distribution. In this way, we can utilize determinative values to estimate the uncertainty of predicted joints coordinates.

Therefore, for the task of UA-HMP, we use the single Gaussian model to measure the uncertainty of predicted joint coordinates. As is shown in Fig \ref{overall_framework}, the output of uncertainty-aware predictor is not determinative joint coordinates but a group of Gaussian parameters. Specifically, we here take $\hat s_t$ as an example: $\hat s_t=\{\hat g_t^i\}_{i=1}^N\in\mathbb{R}^{N\times D}$, where $\hat g_t^i=(\hat\mu_{ x_t^i},\hat\sigma_{ x_t^i},\hat\mu_{ y_t^i},$ $\hat\sigma_{ y_t^i},\hat\mu_{ z_t^i},\hat\sigma_{ z_t^i})$ is the Gaussian parameters of $i$th joint coordinaties at frame $t$. Among these paramaters, the mean value of each coordinate($ie.$ $\hat\mu_{ x_t^i}$) denotes the predicted coordinate of skeleton and each variance value($ie.$ $\hat\sigma_{ x_t^i}$) represents the uncertainty of each coordinate. 

Notably, because only the final predictor layer needs to be modified, our proposed framework can be easily combined with any current baselines to overcome their weakness in uncertainty modeling with slight parameters increment. As a result, our proposed framework has significant practical value for its great generalization on current baselines.

\subsection{Uncertainty-guided learning scheme}
In this part, we illustrate the learning scheme of our proposed UA-HMP. First, considering the uncertainty has no labels to supervise the training phase, quantitating the uncertainty is tricky. Thus, we will explain how to treat and tackle this problem. Second, considering noisy samples can lead to model over-fitting and dramatically degrades the predictive performance, we propose to reduce the negative effect of those noisy samples for better performance.

\subsubsection{Loss Function for uncertainty measurement}

From the perspective of deterministic value, uncertainty measurement is tricky because the uncertainty has no labels in the training phase. However, it can be solved easily from the perspective of distribution. Considering that joint coordinates are output as Gaussian parameters, where the variance represents the uncertainty of each predicted value, we can combine the uncertainty with the predicted value through negative log-likelihood(NLL) loss. We take the Gaussian parameters of ${{ x_t^i}}$ as examples:

\begin{equation}
L_n({x_t^i,\hat \mu_{{x}_t^i},\hat \sigma_{{x}_t^i}})=-log(N(x_t^i|\hat \mu_{{x}_t^i},\hat \sigma_{{x}_t^i}))=-log( {\frac{1}{\sqrt{2\pi\hat\sigma_{{x}_t^i}}}\times exp({-\frac{({x_t^i-\hat\mu_{{x}_t^i}})^2}{2\hat\sigma_{{x}_t^i}}})} )
\label{eq_6}
\end{equation}

Ensentially, equation \ref{eq_6} has a equivalent form:

\begin{equation}
\begin{split}
L_n({x_t^i,\hat \mu_{{x}_t^i},\hat\sigma_{{x}_t^i}})=\frac{1}{2}(log\hat\sigma_{{x}_t^i} +\frac{(x_t^i-\hat\mu_{{x}_t^i})^2}{\hat\sigma_{{x}_t^i}}) + \frac{1}{2}log2\pi
\end{split}
\label{eq_8}
\end{equation}

As stated in the equation \ref{eq_8}, $L_n({x_t^i,\hat \mu_{{x}_t^i},\hat\sigma_{{x}_t^i}})$ consists of two main components (except for the constant term): a regression term divided by the uncertainty and an uncertainty regularization term. During training, the regression term will force the variance $\hat\sigma_{{x}_t^i}$ to get closer to the MSE loss $(x_t^i-\hat\mu_{{x}_t^i})^2$. If the MSE loss gets larger, the variance will get larger. Considering the above situation also means the sample is more unreliable, the variance also represents the prediction uncertainty to some extent. As to the second regularization term, it prevents the network from predicting infinite uncertainty to keep training stability. In this way, we do not need 'uncertainty labels' to learn uncertainty. Instead, it is learned implicitly in the regression task. In brief, $L_n({x_t^i,\hat \mu_{{x}_t^i},\hat\sigma_{{x}_t^i}})$ can quantitate the uncertainty of predicted coordinate of $x_t^i$. Thus, the overall loss function for all predicted motion sequences ${\hat{S}}_{T+1:T_f}$ is as follows:

\begin{equation}
		L_n^U= \frac{1}{N\times(T_f-T)}\sum_{t=T+1}^{T_f}\sum_{i=1}^{N} L_n({x_t^i,\hat \mu_{{x}_t^i},\hat\sigma_{{x}_t^i}}) + L_n({y_t^i,\hat \mu_{{y}_t^i},\hat\sigma_{{y}_t^i}})+ L_n({z_t^i,\hat \mu_{{z}_t^i},\hat\sigma_{{z}_t^i}})
\label{eq_4}
\end{equation}

\begin{table*}[!t] % short-term results on h36m
\caption{Short-term prediction on H$3.6$M. Where ``ms'' denotes ``milliseconds''.}
\scriptsize
\begin{center}
\setlength{\tabcolsep}{1mm}{
\begin{tabular}{c|cccc|cccc|cccc|cccc}%{p{.9cm}p{0.18cm}p{0.18cm}p{0.3cm}p{0.3cm}p{0.3cm}p{0.35cm}|p{0.18cm}p{0.18cm}p{0.3cm}p{0.3cm}p{0.3cm}p{0.35cm}|p{0.18cm}p{0.18cm}p{0.3cm}p{0.3cm}p{0.3cm}p{0.35cm}|p{0.18cm}p{0.18cm}p{0.3cm}p{0.3cm}p{0.3cm}p{0.35cm}}
\hline
motion & \multicolumn{4}{c}{Walking} & \multicolumn{4}{c}{Eating}& \multicolumn{4}{c}{Smoking} & \multicolumn{4}{c}{Discussion}\\
\hline
time(ms)&80&160&320&400&80&160&320&400&80&160&320&400&80&160&320&400 \\
\hline
ResSup \cite{17} &23.8 &40.4& 62.9& 70.9& 17.6& 34.7& 71.9& 87.7& 19.7& 36.6& 61.8& 73.9&31.7& 61.3& 96.0& 103.5 \\
ConvS2S \cite{19} &17.1 &31.2&53.8&61.5&13.7&25.9&52.5&63.3&11.1&21.0&33.4&38.3&18.9&39.3&67.7&75.7\\
% LPJP \cite{32} &7.9 &14.5&29.1&34.5&8.4&18.1&37.4&45.3&6.8&{13.2}&{24.1}&{\bf27.5}&8.3&{21.7}&43.9&48.0\\

\hline
LTD \cite{08}&8.9 &15.7&29.2& 33.4& 8.8& 18.9& 39.4& 47.2& 7.8& 14.9& 25.3&{28.7}& 9.8& 22.1&{\bf39.6} &{\bf44.1} \\

LTD+U &8.9 &16.8&29.1& {\bf33.4}& {\bf8.5}& {\bf17.8}& {\bf36.8}& {\bf44.4}& 7.8& 14.6& 25.2& 29.2& 10.6& 23.2&40.6 &{44.7} \\
\hline
TrajCNN \cite{20} &{\bf8.2} &{\bf14.9}&30.0&35.4&{\bf8.5}&18.4&37.0&44.8&{\bf6.3}&{\bf12.8}&{\bf23.7}&{\bf27.8}&{\bf7.5}&{\bf20.0}&41.3&47.8\\
TrajCNN+U  &8.9 &15.3&{\bf27.7}&{\bf33.4}&9.6&20.9&40.2&48.0&6.5&{13.0}&{24.3}&{29.2}&8.6&{21.9}&43.9&50.6\\

\hline
\hline
motion & \multicolumn{4}{c}{Direction} & \multicolumn{4}{c}{Greeting}& \multicolumn{4}{c}{Phoning} & \multicolumn{4}{c}{Posing}\\
\hline
time(ms)&80&160&320&400&80&160&320&400&80&160&320&400&80&160&320&400 \\
\hline

ResSup \cite{17} & 36.5 &56.4& 81.5& 97.3&37.9& 74.1& 139.0& 158.8 &25.6& 44.4& 74.0& 84.2& 27.9& 54.7& 131.3& 160.8 \\
ConvS2S \cite{19} & 22.0&37.2 &59.6& 73.4 &24.5 &46.2 &90.0& 103.1& 17.2& 29.7& 53.4 &61.3& 16.1& 35.6& 86.2& 105.6\\

% LPJP \cite{32} &11.1 &22.7&48.0&58.4&13.2&28.0&64.5&77.9&10.8&{19.6}&{37.6}&{46.8}&8.3&{22.8}&65.6&81.8\\

\hline
LTD \cite{08}& 12.6 & 24.4&{48.2}&{\bf58.4}& 14.5& 30.5& 74.2& 89.0 & 11.5& 20.2& 37.9& 43.2&9.4& 23.9& {66.2}&{ 82.9}\\

LTD+U &12.0 &{\bf22.3}&48.0 &59.2 & 14.1& 28.3& 69.1& 84.8& 11.6& 19.6& 37.2& {\bf41.6}& 9.2& 22.6& 63.8&80.4  \\

\hline
TrajCNN \cite{20} &{\bf9.7} &{\bf22.3}&50.2&61.7&12.6&28.1&{\bf67.3}&{\bf80.1}&{\bf10.7}&{\bf18.8}&{\bf37.0}&43.1&{\bf6.9}&{\bf21.3}&{\bf62.9}&{\bf78.8}\\
TrajCNN+U  &10.3 &22.4&{\bf47.8}&58.8&{\bf12.5}&{\bf27.5}&69.6&84.1&11.3&{19.6}&{37.9}&{45.1}&7.4&{21.8}&65.3&82.1\\

\hline
\hline{}
motion & \multicolumn{4}{c}{Purchasing} & \multicolumn{4}{c}{Sitting}& \multicolumn{4}{c}{Sitting down} & \multicolumn{4}{c}{Taking photo}\\
\hline
time(ms)&80&160&320&400&80&160&320&400&80&160&320&400&80&160&320&400 \\
\hline
%\hline
ResSup \cite{17} & 40.8& 71.8& 104.2& 109.8 &34.5& 69.9& 126.3 &141.6& 28.6& 55.3& 101.6& 118.9 &23.6 &47.4& 94.0& 112.7\\
ConvS2S \cite{19} & 29.4& 54.9& 82.2& 93.0 &19.8 &42.4& 77.0& 88.4& 17.1& 34.9& 66.3& 77.7& 14.0& 27.2& 53.8& 66.2\\

% LPJP \cite{32} &18.5 &38.1&{\bf61.8}&{\bf69.6}&9.5&23.9&49.8&61.8&11.2&{29.9}&{59.8}&{68.4}&6.3&{14.5}&38.8&49.4\\

\hline
LTD \cite{08}& 19.6&38.5&{64.4}&{\bf72.2}&10.7& 24.6& 50.6&62.0&11.4 &{ 27.6}& 56.4& 67.6& 6.8& 15.2& {38.2}&{49.6} \\
LTD+U &19.4 &38.0&65.0 &74.8 & 10.3& 23.3& 49.5& 61.7& {\bf10.6}& {\bf26.3}& {\bf52.4}& 62.1& 7.1& 14.7& 36.7&48.0  \\

\hline
TrajCNN \cite{20}& {\bf17.1} &{\bf 36.1}&{\bf64.3}&{75.1}& 9.0& 22.0& 49.4& 62.6 & 10.7& 28.8& 55.1& 62.9&{\bf5.4}& {13.4}& {\bf36.2}&{\bf47.0}\\
TrajCNN+U  &18.4 &38.7&65.3&73.9&{\bf8.5}&{\bf20.4}&{\bf43.3}&{\bf54.0}&11.1&{\bf27.4}&{\bf52.9}&{\bf61.0}&5.6&{\bf12.9}&36.8&48.6\\

\hline
\hline
motion & \multicolumn{4}{c}{Waiting} & \multicolumn{4}{c}{Walking dog}& \multicolumn{4}{c}{Walking Together} & \multicolumn{4}{c}{Average}\\
\cline{1-17}%\hline
time(ms)&80&160&320&400&80&160&320&400&80&160&320&400&80&160&320&400 \\
\hline
%\hline
ResSup \cite{17} & 29.5& 60.5& 119.9& 140.6& 60.5& 101.9& 160.8& 188.3& 23.5& 45.0& 71.3& 82.8& 30.8& 57.0& 99.8& 115.5\\
ConvS2S \cite{19} &17.9& 36.5& 74.9& 90.7& 40.6& 74.7& 116.6& 138.7& 15.0& 29.9& 54.3& 65.8& 19.6& 37.8& 68.1& 80.2\\

% LPJP \cite{32} &8.4 &21.5&53.9&69.8&22.9&50.4&100.8&119.8&8.7&{18.3}&{34.2}&{44.1}&10.7&{23.8}&50.0&60.2\\

\hline
LTD \cite{08}& 9.5& 22.0& 57.5& 73.9& 32.2& 58.0& 102.2& 122.7 & 8.9& { 18.4}& 35.3& 44.3& 12.1& 25.0& 51.0& 61.3\\

LTD+U & 9.5& 22.3& 58.9& 76.4& 32.8& 58.3& 100.0& 118.5 & 9.6& { 19.2}& 35.3& 44.5& 12.1& 24.5& 49.8& 60.2\\

\hline
TrajCNN \cite{20}& 8.2 & 21.0&{53.4}&{68.9}& 23.6& 52.0& 98.1& 116.9 & 8.5& 18.5& 33.9& 43.4&{\bf10.2}& 23.2& {49.3}&{ 59.7}\\
TrajCNN+U  &{\bf8.1} &{\bf19.4}&{\bf49.8}&{\bf65.7}&{\bf22.7}&{\bf50.0}&{\bf96.1}&{\bf115.8}&{\bf8.2}&{\bf17.4}&{\bf33.5}&{\bf42.5}&10.5&{\bf23.2}&{\bf49.0}&{\bf59.5}\\
\hline

\end{tabular}
}

\end{center}
\label{r_h36mshort}
\vspace{-2.5em}
\end{table*}

\subsubsection{Loss Function for uncertainty guidance}

It is necessary to reduce the negative effect of noisy samples because noisy samples can lead to model over-fitting and dramatically degrades the predictive performance. Considering the resulting uncertainty in Section 3.3.1 represents the reliability of prediction, this parameter can be used as a penalty coefficient to raw loss function MPJPE used in most current methods. In particular, for one training sample, the MPJPE loss $L_m$ is as follows:
\begin{equation}{}
	\begin{split}
		L_m &= \frac{1}{N\times(T_f-T)}\sum_{t=T+1}^{T_f}\sum_{i=1}^{N} L_p({\hat{p}}_t^i,{p_t^i})
	\end{split}
\label{eq_4}
\end{equation}

where $L_p({\hat{p}}_t^i,{p_t^i})$ represents the 2 norm of predicted joint ${p_t^i}$${\hat{p}}_t^i$ and corresponding ground truth ${p_t^i}$. After combining the penalty coefficient to the $L_m$, the resulting loss function is as follows:

\begin{equation}
	\begin{split}
		L_m^U &= \frac{1}{N\times(T_f-T)}\sum_{t=T+1}^{T_f}\sum_{i=1}^{N} L_p({\hat{p}}_t^i,{p_t^i}) * w(\hat\sigma_{{x}_t^i},\hat\sigma_{{y}_t^i},\hat\sigma_{{x}_t^i})
	\end{split}
\label{eq_4}
\end{equation}

where $w(\hat\sigma_{{x}_t^i},\hat\sigma_{{y}_t^i},\hat\sigma_{{x}_t^i})=\frac{1}3\times(
(\hat\sigma_{{x}_t^i})^k+(\hat\sigma_{{y}_t^i})^k+(\hat\sigma_{{z}_t^i})^k)$ represents the averaged penalty weights of joint $i$ at time $t$. k is a temperature coefficient and is set as -0.2 in our experiments). More details can be found in supplementary materials.

\subsubsection{Overall Loss Function}

In brief, the final loss function of predicted motion sequences ${\hat{S}}_{T+1:T_f}$ is as follows: $L^U = L_m^U + L_n^U$, where $L_m^U$ is used to measure uncertainty and $L_n^U$ is proposed to eliminate the negative effect of noisy samples with high uncertainty for better optimization during training.

\begin{table*}[!t] % results on cmu
\caption{Short and long-term prediction on CMU-mocap.}
\scriptsize
\begin{center}
\setlength{\tabcolsep}{1mm}{
\begin{tabular}{c|ccccc|ccccc|ccccc}%{p{.9cm}p{0.18cm}p{0.18cm}p{0.3cm}p{0.3cm}p{0.35cm}|p{0.18cm}p{0.18cm}p{0.3cm}p{0.3cm}p{0.35cm}|p{0.18cm}p{0.18cm}p{0.3cm}p{0.3cm}p{0.35cm}
\hline%%%%%%%%%%%
motion& \multicolumn{5}{c}{Basketball} & \multicolumn{5}{c}{Basketball Signal}& \multicolumn{5}{c}{Directing Traffic}\\
\hline
time (ms) & 80 &160 & 320 &400 &1000& 80 &160 & 320 &400 &1000& 80 &160 & 320 &400 &1000\\
\hline
LTD \cite{08}&14.0&25.4 &  49.6 &61.4&{104.79}   &3.5 &   6.1 & 11.7 &15.2 &   45.1& 7.4 & 15.1 &31.7 &   42.2 &142.1\\

LTD+U &12.3&22.0 &  44.0 &657.0&{106.1}   &2.5 &   4.4 & 10.7 &14.9 &   53.9& 5.7 & 11.1 &24.5 &   31.3 &152.4\\

\hline
TrajCNN \cite{20}&11.1&19.7   &43.9&   {\bf56.8}& {114.1}&   {1.8} &3.5  &{9.1}   &13.0 &{49.6}  &{\bf5.5} & {\bf10.9}   &{\bf23.7} &{\bf31.3}&  {\bf105.9}\\

TrajCNN+U &{\bf10.4}&{\bf18.4}   &{\bf43.5}&   57.1& {\bf110.7}&   {\bf1.7} &{\bf3.3}  &{\bf8.1}   &{\bf11.7} &{\bf47.1}  &{5.6} & {11.6}   &{24.6} &{33.3}&  {132.0}\\

\hline%%%%%%%%%%%%%%%%
motion& \multicolumn{5}{c}{Jumping} & \multicolumn{5}{c}{Running}& \multicolumn{5}{c}{Soccer}\\
\hline
time (ms) & 80 &160 & 320 &400 &1000& 80 &160 & 320 &400 &1000& 80 &160 & 320 &400 &1000\\
\hline
LTD \cite{08}&16.9&  34.4 &   76.3  &96.8    &{\bf164.6}   &25.5    &36.7&   39.3 &   39.9  &58.2 &11.3    &21.5    &44.2&   55.8 &   117.5\\

LTD+U &14.2&30.0 &  {\bf71.6} &{\bf92.9}&{175.6}   &17.6 &   21.5 & 21.9 &27.8 &   69.5& 9.9 & 18.7 &38.8 &   49.4 &114.3\\
\hline
TrajCNN \cite{20}&{\bf12.2}  &28.8&   {72.1}&  94.6& 166.0 &17.1&   24.4  &28.4&   32.8& {49.2}  &{8.1}   &{17.6}& 40.9  &51.3 &126.5\\

TrajCNN+U &{\bf12.2}&{\bf27.8}   &73.8&   99.1& 173.2&   {\bf17.0} &{\bf19.8}  &{\bf19.8}   &{\bf26.1} &{\bf42.9}  &{\bf6.0} & {\bf7.8}   &{\bf16.3} &{\bf35.8}&  {\bf100.8}\\
\hline

\end{tabular}
}
\begin{tabular}{c|ccccc|ccccc|ccccc}%{p{.9cm}p{0.18cm}p{0.18cm}p{0.3cm}p{0.3cm}p{0.35cm}|%{c|ccccc|ccccc|ccccc|c}
\hline%%%%%%%%%%%
motion& \multicolumn{5}{c}{Walking} & \multicolumn{5}{c}{Wash Window}& \multicolumn{5}{c}{Average}\\
 \cline{1-16}
time (ms) & 80 &160 & 320 &400 &1000& 80 &160 & 320 &400 &1000& 80 &160 & 320 &400 &1000\\
\hline
LTD \cite{08}&7.7 &  11.8 &   19.4 &   23.1 &   40.2  &5.9 &   11.9 &   30.3 &   40.0 &   79.3  &11.5 &  20.4 &   37.8 &   46.8 &   96.5\\
LTD+U &{\bf5.9} &  {\bf9.6} &   {\bf17.3} &   {\bf21.2} &   {\bf37.4} &4.9 &   10.8 &   29.3 &  38.6 &   {\bf76.6}  &9.1 &  15.9 &   32.3 &   41.6 &   95.7\\
\hline
TrajCNN \cite{20}&6.5   &10.3&   19.4& 23.7& 41.6  &{4.5}&{9.7}  &29.9&   41.5& 89.9  &8.3  &15.6&   33.4  &43.1 &{92.8}\\

TrajCNN+U &6.7&10.9   &18.9&   23.1& 40.8&   {\bf4.0} &{\bf8.3} &{\bf26.2}   &{\bf37.4} &{\bf81.9}  &{\bf8.1} & {\bf14.5}   &{\bf31.3} &{\bf41.8}&  {\bf91.2}\\
\hline
\end{tabular}
\end{center}
\label{results_cmu}
\vspace{-2.5em}
\end{table*}

\begin{table*}[!t] % long-term results on h36m
\vspace{0.5em}
\caption{Long-term prediction on H$3.6$M.}
\scriptsize
\begin{center}
\setlength{\tabcolsep}{1mm}{
\begin{tabular}{c|cc|cc|cc|cc|cc|cc|cc|cc}%{p{.9cm}p{0.18cm}p{0.18cm}p{0.3cm}p{0.3cm}p{0.3cm}p{0.35cm}|p{0.18cm}p{0.18cm}p{0.3cm}p{0.3cm}p{0.3cm}p{0.35cm}|p{0.18cm}p{0.18cm}p{0.3cm}p{0.3cm}p{0.3cm}p{0.35cm}|p{0.18cm}p{0.18cm}p{0.3cm}p{0.3cm}p{0.3cm}p{0.35cm}}
\hline
motion & \multicolumn{2}{c}{Walking} & \multicolumn{2}{c}{Eating}& \multicolumn{2}{c}{Smoking} & \multicolumn{2}{c}{Discussion}
&\multicolumn{2}{c}{Directions} & \multicolumn{2}{c}{Greeting}& \multicolumn{2}{c}{Phoning} & \multicolumn{2}{c}{Posing}\\
\hline
time(ms)&560 &1000&560 &1000&560 &1000&560 &1000&560 &1000&560 &1000&560 &1000&560 &1000\\
\hline
LTD \cite{08}&42.2&51.3&{\bf56.5}&{\bf68.6}&{\bf32.3}&60.5&{70.4}&103.5&85.8&109.3&91.8&87.4&65.0&113.6&113.4&220.6\\
LTD+U &41.8&45.1&{59.2}&{70.1}&33.0&62.1&{\bf67.4}&102.4&80.6&102.0&97.7&90.9&64.2&113.8&105.4&{\bf207.8}\\
\hline
TrajCNN \cite{20}&{\bf37.9}&46.4&59.2&71.5&32.7&{\bf58.7}&75.4&{103.0}&{84.7}&104.2&{\bf91.4}&{\bf84.3}&62.3&113.5&111.6&{210.9}\\
TrajCNN+U&39.4&{\bf45.1}&57.4&73.4&33.8&59.9&70.9&{\bf97.7}&{\bf72.9}&98.9&98.8&{89.3}&{\bf56.5}&{\bf107.1}&{\bf102.8}&{209.0}\\
\hline
\hline
\end{tabular}
}
\setlength{\tabcolsep}{1mm}{
\begin{tabular}{c|cc|cc|cc|cc|cc|cc|cc|cc}
\hline
motion & \multicolumn{2}{c}{Purchases} & \multicolumn{2}{c}{Sitting}& \multicolumn{2}{c}{Sitting down} & \multicolumn{2}{c}{Taking photo}
&\multicolumn{2}{c}{Waiting} & \multicolumn{2}{c}{Walking Dog}& \multicolumn{2}{c}{Walking Tog} &\multicolumn{2}{c}{Average}\\
\cline{1-17}
time(ms)&560 &1000&560 &1000&560 &1000&560 &1000&560 &1000&560 &1000&560 &1000&560 &1000\\
\hline
LTD \cite{08}&94.3&130.4&79.6&114.9&82.6&140.1&68.9&87.1&100.9&167.6&136.6 &174.3&{\bf57.0}&85.0&78.5&{\bf114.3}\\
LTD+U &89.6&124.8&80.1&114.3&83.5&130.5&73.1&87.7&103.0&167.1&143.4 &177.1&{58.4}&81.1&78.7&111.8\\
\hline
TrajCNN \cite{20}&84.5&{\bf115.5}&81.0&116.3&{\bf79.8}&{123.8}&{\bf73.0}&{\bf86.6}&{\bf92.9}&{\bf165.9}&141.1&181.3&57.6&{\bf77.3}&77.7&110.6\\
TrajCNN+U &{\bf82.5}&{123.9}&{\bf78.0}&116.9&81.6&{\bf117.9}&{74.9}&{87.9}&95.1&172.0&{\bf129.6}&{\bf172.5}&64.6&{83.8}&{\bf75.8}&{\bf110.4}\\
\hline
\end{tabular}
}
\end{center}
\label{r_h36mlong}
\vspace{-2.0em}
\end{table*}

\section{Experiments}
We evaluate our model on several benchmark motion capture (mocap) datasets, including Human3.6M (H3.6M) \cite{46} and the CMU mocap dataset. We first introduce some experimental details. Next, we will demonstrate the performance of our framework quantitatively and qualitatively.

\subsection{Experimental details}

\subsubsection{Datasets}

\textbf{H3.6M.} \cite{46} is the most widely used benchmark for motion prediction. It involves 15 actions and each human pose involves a 32-joint skeleton. To remove the global rotation, translation, and constant 3D coordinates of each human pose, there remain 22 joints. 

% We test our method on subject 5 (S5).
% Following \cite{20,08}, we compute the joint's 3D coordinates by applying forward kinematics and down-sample the motion sequence to 25 frames per second.
\begin{figure}[h]
\centering 
\includegraphics[width=3.5in]{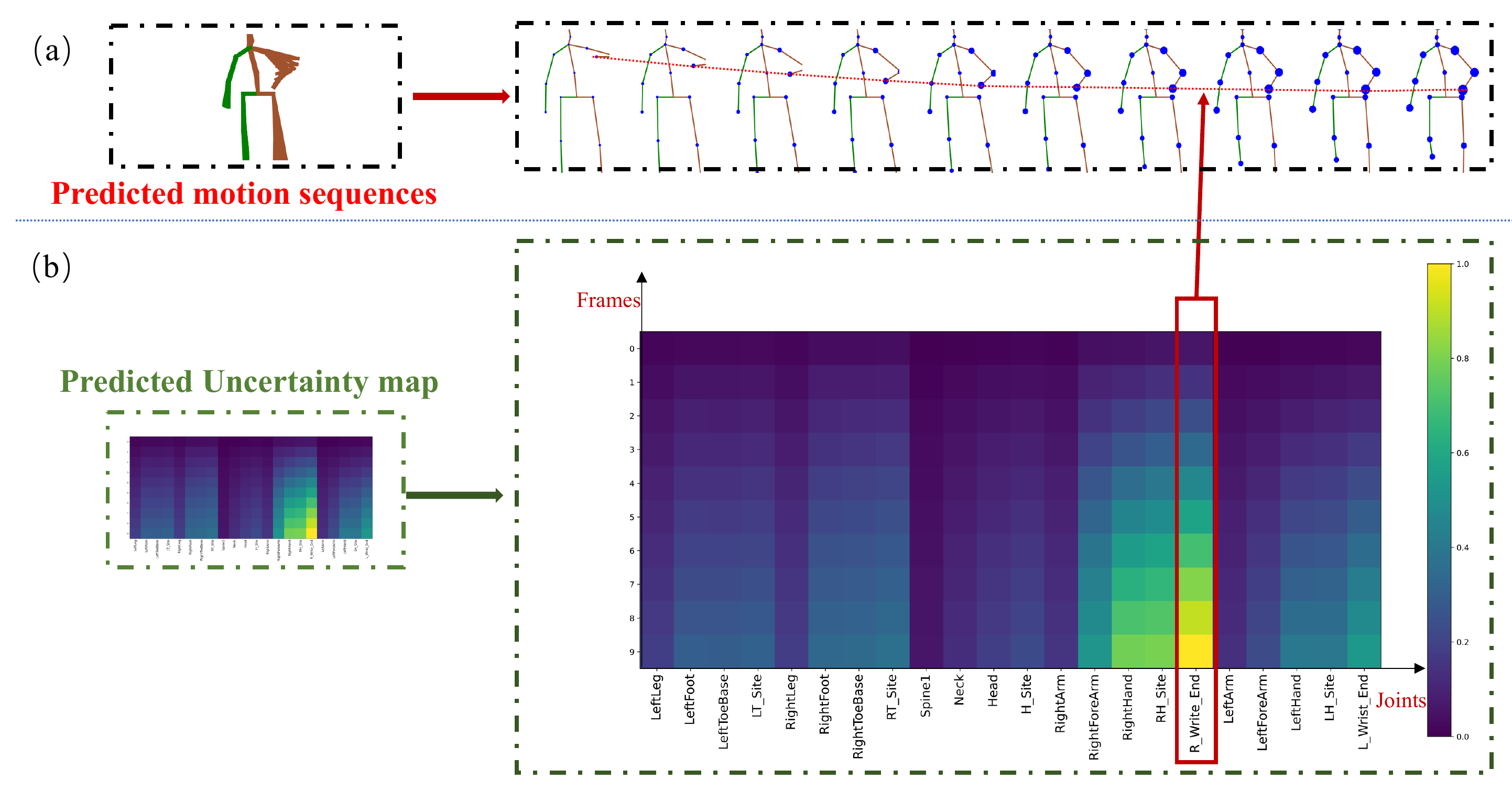} 
\caption{Two visualizations of uncertainty. (a) The size of the point (the larger size represents the larger uncertainty). (b) The brightness of the uncertainty map (the brighter of the element, the large uncertainty of its corresponding joint). The red box and red line illustrate the corresponding relationships between the two visualizations. }
\label{Vis_details}
\end{figure}

\textbf{CMU-Mocap.} The CMU mocap dataset mainly includes five categories. Be consistent with \cite{08,20}, we select 8 detailed actions: ``basketball'', ``basketball signal'', ``directing traffic'', ``jumping'', ``running'', ``soccer'', ``walking'' and ``washing window''.

\subsubsection{Baselines and implementation details}
We combine our framework with two types of baselines: LTD\cite{08} and TrajCNN\cite{20}.

\textbf{LTD}\cite{08}: A GCN-based deep network for motion prediction, which takes into account both temporal smoothness and spatial dependencies among human body joints.

\textbf{TrajCNN}\cite{20}: A CNN-based deep network designed for modeling motion dynamics of the input sequence with coupled spatio-temporal features, dynamic local-global features, and global temporal co-occurrence features in the new space.

All the training settings of the experiments are following the raw baselines.

\subsection{Qualitative results}
% \subsubsection{Comparison with state-of-the-art}
Here we show the prediction performance for short-term and long-term motion prediction on H3.6M, CMU Mocap. We quantitatively evaluate various methods by the MPJPE between the generated motions and ground truths in 3D coordinates space. Limited by the pages, we will provide more ablation study in supplementary materials. ('+U' means combining the raw basslines with our UA-HMP.)

% To be consistent with the literature\cite{20,08}, we report our results for short-term ($<$ 500ms) and long-term ($>$ 500ms) predictions. For all datasets, we are given ten frames (400 milliseconds) to predict the future ten frames (400 milliseconds) for short-term prediction and to predict the future 25 frames (1 second) for long-term prediction. More results can be found in the supplementary material.

\textbf{Short-term motion prediction on H3.6M.} Table \ref{r_h36mshort} provides the short-term predictions on H3.6M for the 15 activities and the average results. Note that those baselines combined with our UA-HMP outperform all raw baselines on average and almost all motions. It demonstrates the effectiveness of our processed uncertainty-guided learning scheme. By reducing the negative effect of the noisy samples, the performance gets better, especially on 320ms and 400ms. This shows our method is more robust faced with the varying of time. 
 
\textbf{Long-term motion prediction on H3.6M.} In Table \ref{r_h36mlong}, we compare our results with those of the baselines for long-term prediction on H3.6M. Our method outperforms almost the baselines on average. For long-term prediction, our method still obtains competitive performances on almost all motions with the uncertainly of motion increasing.

\begin{figure}[h]
\centering 
\includegraphics[width=4.5in]{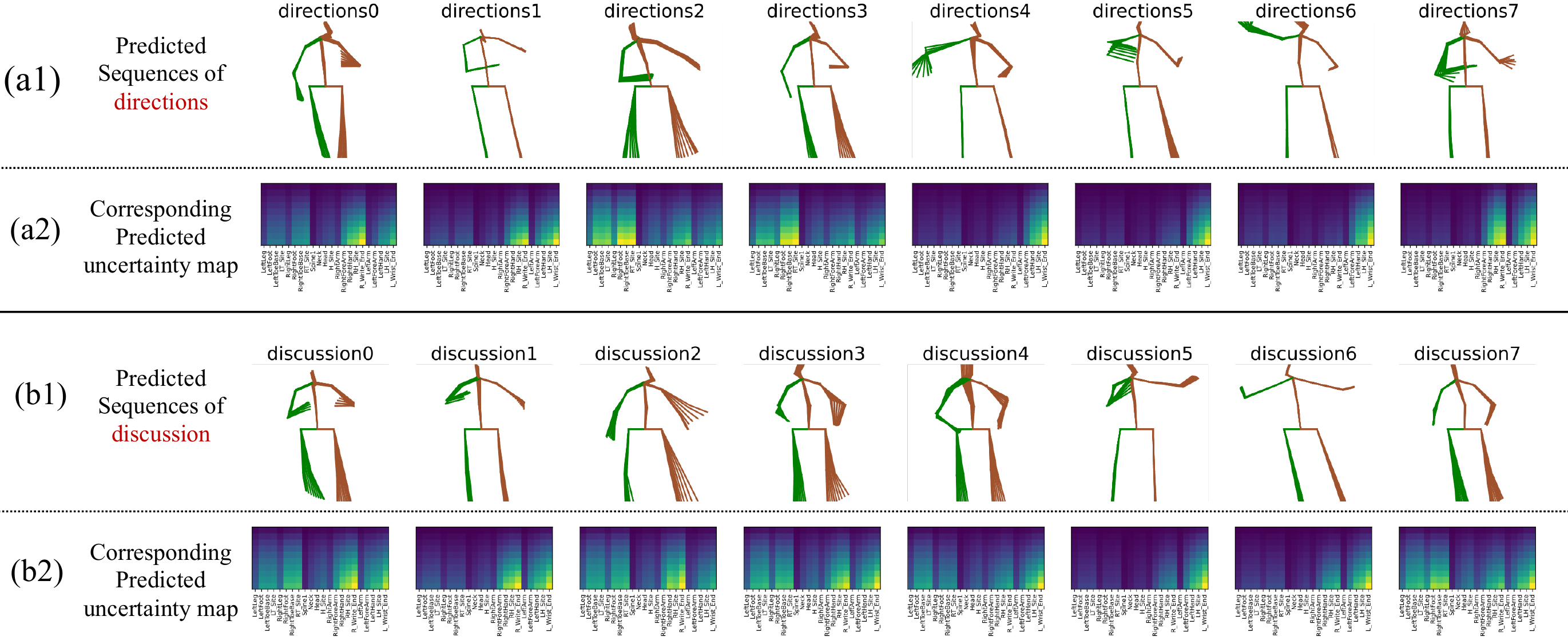} 
\caption{More Visualizations. The (a1),(b1) are predicted sequences of different motion sequences. They show the stacked motion sequences by stacking multiple skeletons together. The (a2),(b2) show the uncertainty map of the corresponding motion sequences, where the horizontal axis represents joints, and the vertical axis represents time.  }
\label{Vis_all}
\end{figure}

\textbf{Prediction on CMU-Mocap} 
Table \ref{results_cmu} reports the results on CMU-Mocap. Our framework combined with existing baselines outperforms the raw baselines for both short-term and long-term prediction. Notably, the degree of improvement in CMU-Mocap is larger than H3.6M, demonstrating that our proposed uncertainty-guided learning scheme is more beneficial for situations without large training samples.

\subsection{Quantitative results}
% As is shown in Fig.\ref{Vis_details}, it shows the relationships between our predicted motion sequences and uncertainty map. The blue point on the right represents the degree of uncertainty. With the degree of uncertainty larger, the point gets larger and the uncertainty map gets brighter. It explicitly shows that in this action "greeting", the right leg movement is the largest, and the corresponding uncertainty value is the largest. More visualization on different motions is also provided in Fig\ref{Vis_all}.

We show the uncertainty by two methods in Figure \ref{Vis_details}. In (a), the left part is the stacked motion sequences, and the right part is the details of the left one. Here, we use the size of the point to represent uncertainty. The larger size represents the larger uncertainty. We can easily see the uncertainty of every prediction by this illustration. From the horizontal change of the elements, we can see the uncertainty of the point is larger with the time duration increasing. In (b), the uncertainty map is used to illustrate uncertainty. The brighter the element in the map, the larger uncertainty of its corresponding joint. By this illustration, we can easily read the uncertainty evolution from the temporal dimension and the joint dimension separately. From the horizontal change of the elements, we can see different uncertainty of different joints. More visualizations are demonstrated in Figure \ref{Vis_all} and supplementary materials.

\section{Conclusion}

In this paper, we present our uncertainty-aware framework for human motion prediction (UA-HMP). It mainly includes two core components. First, we design an uncertainty-aware predictor through Gaussian modeling to get the value and the uncertainty of predicted motion. Second, we present an uncertainty-guided learning scheme to quantitate the uncertainty and improve prediction accuracy. In particular, those samples with high uncertainty are given low weight during optimization. In this way, the adverse effect of unreliable samples can be avoided for better optimization in the training phase. Our proposed framework is easily combined with current SOTA baselines to overcome their weakness in uncertainty modeling. As a result, our proposed framework has significant practical value.

\bibliography{egbib}
\end{document}